\documentclass{article}

\usepackage[preprint]{neurips_2025}

\usepackage[utf8]{inputenc}
\usepackage[T1]{fontenc}
\usepackage{hyperref}
\usepackage{url}
\usepackage{booktabs}
\usepackage{amsfonts}
\usepackage{amsmath}
\usepackage{amssymb}
\usepackage{nicefrac}
\usepackage{microtype}
\usepackage{xcolor}
\usepackage{graphicx}
\usepackage{algorithm}
\usepackage{algorithmic}
\usepackage{enumitem}
\usepackage{multirow}
\usepackage{array}
\usepackage{tabularx}
\usepackage{listings}

\definecolor{codegreen}{rgb}{0,0.6,0}
\definecolor{codegray}{rgb}{0.5,0.5,0.5}
\definecolor{codepurple}{rgb}{0.58,0,0.82}
\definecolor{backcolour}{rgb}{0.95,0.95,0.92}

\lstdefinestyle{mystyle}{
    backgroundcolor=\color{backcolour},   
    commentstyle=\color{codegreen},
    keywordstyle=\color{magenta},
    numberstyle=\tiny\color{codegray},
    stringstyle=\color{codepurple},
    basicstyle=\ttfamily\footnotesize,
    breakatwhitespace=false,         
    breaklines=true,                 
    captionpos=b,                    
    keepspaces=true,                 
    numbers=left,                    
    numbersep=5pt,                  
    showspaces=false,                
    showstringspaces=false,
    showtabs=false,                  
    tabsize=2
}

\title{PKG-DPO: Optimizing Domain-Specific AI systems with Physics Knowledge Graphs and Direct Preference Optimization }

\author{%
  Nitin Nagesh Kulkarni* \\
  Advanced Engineering and Technology \\
  Milwaukee Tool \\
  Brookfield, WI 53005, USA \\
  \And
  Bryson Wilcox \\
  Advanced Engineering and Technology \\
  Milwaukee Tool \\
  Brookfield, WI 53005, USA \\
  \And
  Max Sawa \\
  Advanced Engineering and Technology \\
  Milwaukee Tool \\
  Brookfield, WI 53005, USA \\
  \And
  Jason Thom \\
  Advanced Engineering and Technology \\
  Milwaukee Tool \\
  Brookfield, WI 53005, USA \\
}

\begin{document}

\maketitle

\begin{abstract}
Advancing AI systems in scientific domains like physics, materials science, and engineering calls for reasoning over complex, multi-physics phenomena while respecting governing principles. Although Large Language Models (LLMs) and existing preference optimization techniques perform well on standard benchmarks, they often struggle to differentiate between physically valid and invalid reasoning. This shortcoming becomes critical in high-stakes applications like metal joining, where seemingly plausible yet physically incorrect recommendations can lead to defects, material waste, equipment damage, and serious safety risks. To address this challenge, we introduce PKG-DPO, a novel framework that integrates Physics Knowledge Graphs (PKGs) with Direct Preference Optimization (DPO) to enforce physical validity in AI-generated outputs. PKG-DPO comprises three key components A) hierarchical physics knowledge graph that encodes cross-domain relationships, conservation laws, and thermodynamic principles. B) A physics reasoning engine that leverages structured knowledge to improve discrimination between physically consistent and inconsistent responses. C) A physics-grounded evaluation suite designed to assess compliance with domain-specific constraints. PKG-DPO achieves 17\% fewer constraint violations and an 11\% higher Physics Score compared to KG-DPO (knowledge graph-based DPO). Additionally, PKG-DPO demonstrates a 12\% higher relevant parameter accuracy and a 7\% higher quality alignment in reasoning accuracy. While our primary focus is on metal joining, the framework is broadly applicable to other multi-scale, physics-driven domains, offering a principled approach to embedding scientific constraints into preference learning.
\end{abstract}

\section{Introduction}

Deploying Large Language Models (LLMs) in domains governed by physical laws introduces challenges that go beyond traditional preference alignment [1]. While Direct Preference Optimization (DPO) has shown promise in aligning models with human preferences, it lacks mechanisms to enforce compliance with physical constraints. This limitation is particularly critical in high-stakes fields such as materials science, engineering, and manufacturing, where violations of physical laws can lead to structural failures, safety hazards, and economic losses [2, 3]. For example, in welding engineering, AI-generated recommendations must satisfy thermodynamic constraints, electrical safety limits, and metallurgical principles simultaneously. Existing preference learning methods often optimize for human-perceived quality without validating physical feasibility. As a result, models may produce outputs that appear plausible but suggest parameters like sub-melting-point temperatures or excessive current densities, which are physically invalid and potentially dangerous. This disconnect is especially pronounced in multi-physics environments, where thermal, electrical, mechanical, and metallurgical interactions must be considered concurrently. Traditional preference learning lacks the structured domain knowledge required to validate outputs against conservation laws and safety thresholds, leading to fluent but physically incorrect recommendations [4].

Efforts to integrate physical constraints into machine learning have led to the development of Physics-Informed Neural Networks (PINNs), which embed physical laws into loss functions to ensure consistent predictions [5, 6]. These methods are effective for solving differential equations but are not designed for discrete preference learning. Knowledge graphs have emerged as tools for representing domain-specific relationships, enabling constraint modeling and quality assessment in engineering applications [7]. Recent work has demonstrated the utility of dynamically updatable knowledge graphs in Computer-Aided Process Planning (CAPP), facilitating complex reasoning over engineering data [8]. Graph Neural Networks (GNNs) enhance learning over graph-structured data through message passing and multi-hop reasoning, uncovering hidden relationships in biomedical and engineering domains [9, 10]. These techniques have been applied to tasks such as drug repurpose and supply chain risk management, demonstrating their versatility and effectiveness [11, 12]. Despite progress in foundational areas, several gaps hinder the effective use of LLMs in physics-constrained domains:

\begin{enumerate}[noitemsep]
\item \textbf {Domain Knowledge Integration}: DPO struggles to reconcile pre-trained models with domain-specific constraints, especially when expert knowledge contradicts human preferences
\item \textbf {Multi-Objective Alignment}: Optimizing multiple objectives can lead to misalignment or model collapse, complicating iterative fine-tuning.
\item \textbf{Safety-Critical Validation}: Detecting and penalizing responses that could lead to hazardous conditions
\end{enumerate}

\noindent We propose PKG-DPO, a framework that integrates Physics Knowledge Graphs (PKGs) with DPO to enforce physical validity in AI-generated outputs. The framework introduces three key novelties:
\begin{itemize}[noitemsep]
\item \textbf{Hierarchical Physics Knowledge Graph}: Encodes cross-domain relationships, conservation laws, and thermodynamic principles for systematic constraint validation.
\item \textbf{Physics-Aware Preference Optimization}: Enhances DPO by leveraging structured domain knowledge to distinguish physically valid responses
\item \textbf{Domain-Constrained Evaluation Framework}: Assesses compliance with physical constraints across multiple validation dimensions
\end{itemize}

\section{Methodology}
The proposed PKG-DPO framework consists of three sequential stages, as shown in Figure 1. In the first stage, a Physics Knowledge Graph (PKG) is constructed to systematically represent selected physics concepts, governing equations, constraints, and their interdependencies in a structured, machine-readable format. This serves as a persistent, interpretable knowledge base. The second stage involves the development of a Physics Reasoning Engine that leverages the PKG to perform domain-specific inference, enforce physical laws, and validate outputs through built-in safety and consistency checks. This ensures that any generated or inferred results remain within the bounds of scientifically plausible behavior. In the third stage, this physics-grounded reasoning is integrated into AI models using a weighted Direct Preference Optimization (DPO) approach, which strategically balances general performance metrics with physics-specific accuracy. This three-stage pipeline enables AI systems to maintain high task performance while adhering to fundamental physical principles.

\begin{figure}[htbp]
    \centering
    \includegraphics[width=3 in]{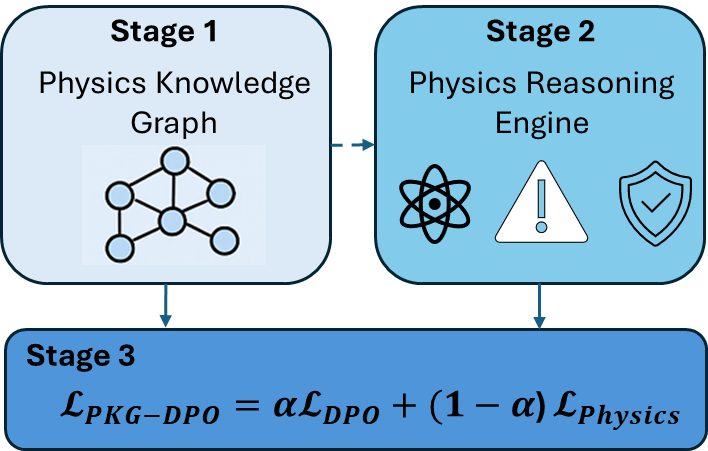} 
    \caption{The overview of PKG-DPO process}
    \label{fig:pkg_dpo}
\end{figure}

\subsection{Physics Knowledge Graph Construction}

The proposed methodology initiates with the development of a comprehensive Physics Knowledge Graph (PKG), designed to encode domain specific (for our case welding process) entities, relationships, and constraints derived from established physical principles and empirical process knowledge.

\textbf{Entities}: Materials (aluminum, steel), processes (GTAW, GMAW), parameters (current, voltage, temperature), constraints (absolute zero, safety limits), and outcomes (defects, quality measures).

\textbf{Relations}: We define a rich set of relation types:
\begin{itemize}[noitemsep]
\item \texttt{CAUSES}: Causal relationships (high current $\rightarrow$ increased penetration)
\item \texttt{PREVENTS}: Preventive relationships (proper cleaning $\rightarrow$ prevents porosity)
\item \texttt{REQUIRES}: Dependency relationships (aluminum welding $\rightarrow$ requires AC current)
\item \texttt{INCOMPATIBLE\_WITH}: Conflict relationships (high speed $\rightarrow$ incompatible with thick sections)
\item \texttt{RANGES}: Quantitative bounds (GTAW current $\in [5\text{A}, 500\text{A}]$)
\end{itemize}

\textbf{Constraints}: Fundamental physics constraints including:
\begin{itemize}[noitemsep]
\item Thermodynamic limits: $T > -273.15°\text{C}$ (absolute zero)
\item Electrical safety: $I > 0\text{A}$, $V > 0\text{V}$
\item Conservation laws: $\eta \leq 1.0$ (efficiency cannot exceed 100\%)
\item Process-specific bounds: $H = \frac{I \times V \times \eta}{v}$ (heat input formula)
\end{itemize}
where $I$ is current, $V$ is voltage, $\eta$ is efficiency and $v$ is travelling speed. This structured representation enables the PKG to serve as a knowledge repository for next stages.

\subsection{Physics Reasoning Engine}

The first approach employs multi-hop traversal of the PKG to identify reasoning paths between source entities and target concepts. A breadth-first search (BFS) algorithm is used to systematically explore relational connections, enabling transparent tracing of causal or dependency chains

\textbf{Step 1: Direct Graph Traversal}
 Algorithm 1 is used to construct reasoning paths between source entities and target concepts by exploring a physics knowledge graph through breadth-first search up to a specified depth. It identifies valid reasoning chains by checking relational consistency at each step and accumulates paths that connect source entities to target concepts while enforcing physical relevance.

\begin{algorithm}
\caption{Multi-Hop Physics Reasoning}
\begin{algorithmic}
\STATE \textbf{Input:} Source entities $S$, target concepts $T$, max depth $d$
\STATE \textbf{Output:} Reasoning paths $\mathcal{P}$
\STATE Initialize queue $Q \leftarrow [(s, [s]) \text{ for } s \in S]$
\STATE Initialize paths $\mathcal{P} \leftarrow \emptyset$
\WHILE{$Q \neq \emptyset$ and $|\mathcal{P}| < 10$}
    \STATE $(current, path) \leftarrow Q.pop()$
    \IF{$len(path) > d$}
        \STATE continue
    \ENDIF
    \FOR{$target \in T$}
        \IF{$target$ relates to $current$}
            \STATE $\mathcal{P} \leftarrow \mathcal{P} \cup \{path\}$
        \ENDIF
    \ENDFOR
    \FOR{$neighbor \in neighbors(current)$}
        \STATE $Q.append((neighbor, path + [neighbor]))$
    \ENDFOR
\ENDWHILE
\RETURN $\mathcal{P}$
\end{algorithmic}
\end{algorithm}

\textbf{Step 2: Constraint-Based Inference}
In this step, candidate reasoning outcomes are validated against formal physics constraints embedded in the PKG. These include thermodynamic limits, electrical safety thresholds, conservation laws, and process-specific equations. Any reasoning path that violates a constraint is pruned, ensuring physical plausibility in all inferred outcomes

\textbf{Step 3: Quantitative Relationship Validation}
The third step evaluates inferred relationships against established quantitative process models. For instance, heat input in welding processes is validated using the equation:
\begin{equation}
\text{Heat Input} = \frac{\text{Current} \times \text{Voltage} \times \text{Efficiency}}{\text{Travel Speed}}
\end{equation}
By cross-verifying numerical predictions against the governing equations, the framework ensures numerical consistency in addition to relational and constraint validity.

\subsection{PKG-DPO Objective Function}

To integrate physics-based reasoning into the Direct Preference Optimization (DPO) framework, we modify the standard objective to jointly optimize for human preference alignment and physics compliance. The proposed PKG-DPO objective is defined as:

\begin{equation}
\mathcal{L}_{PKG-DPO} = \alpha \mathcal{L}_{DPO} + (1-\alpha) \mathcal{L}_{PKG}
\end{equation}

where $\alpha$ balances preference learning and physics compliance, $\mathcal{L}_{DPO}$ is DPO loss and $\mathcal{L}_{PKG}$ is Loss in physics knowledge graph and:

\begin{equation}
\mathcal{L}_{PKG} = \mathbb{E}_{(x,y) \sim \mathcal{D}} \left[ \lambda_1 V(y) + \lambda_2 (1-C(y)) + \lambda_3 (1-R(y)) \right]
\end{equation}

\textbf{Violation Penalty} $V(y)$: Penalizes responses containing physics violations
\begin{equation}
V(y) = \sum_{v \in \text{violations}(y)} w_v \cdot s_v
\end{equation}
where $w_v$ and $s_v$ are weights and severity scores for violation $v$.

\textbf{Coverage Reward} $C(y)$: Rewards responses demonstrating domain knowledge
\begin{equation}
C(y) = \frac{|\text{entities}(y) \cap \text{PKG}|}{|\text{entities}(y)|}
\end{equation}

\textbf{Reasoning Reward} $R(y)$: Rewards responses following valid reasoning paths
\begin{equation}
R(y) = \frac{1}{|\text{paths}|} \sum_{p \in \text{paths}(y)} \text{confidence}(p)
\end{equation}

\subsection{Enhanced Preference Data Processing}

In PKG-DPO, conventional preference pairs $(x, y_w, y_l)$  where $y_w$ denotes the preferred response and $y_l$ the less preferred one are augmented into enriched tuples:

\begin{equation}
(x, y_w, y_l, \mathcal{V}_w, \mathcal{V}_l, \mathcal{P}_w, \mathcal{P}_l, s_w^{\mathrm{PKG}}, s_l^{\mathrm{PKG}})
\end{equation}

Here:
\begin{itemize}[noitemsep]
    \item $\mathcal{V}_w, \mathcal{V}_l$: Quantified physics violations in the preferred and rejected responses, respectively.
    \item $\mathcal{P}_w, \mathcal{P}_l$: Physics-informed reasoning paths derived from the PKG for each response.
    \item $s_w^{\mathrm{PKG}}, s_l^{\mathrm{PKG}}$: Physics consistency scores computed by the PKG-based evaluation engine.
\end{itemize}

The above carries enriched representation captures both the linguistic preference signal and explicit physics-grounded validation, enabling the optimization process to jointly consider task performance and scientific plausibility.

\section{Experimental Setup}

The experimental framework for evaluating physics-constrained preference learning in welding technical knowledge systems was designed to rigorously assess both domain expertise and adherence to fundamental physical principles. It consists of three core components: dataset construction with expert annotation, baseline method implementation, and multi-dimensional performance evaluation.

We used Phi-3-mini-4k-instruct as the backbone model for all preference learning experiments [14]. This lightweight yet capable LLM was selected for its efficiency and adaptability in domain-specific reasoning tasks. The dataset comprises over 10{,}000 expert-validated preference pairs covering the full spectrum of modern welding practices—equipment setup, process principles, defect analysis, metallurgy, safety protocols, and advanced techniques. Each pair reflects varying levels of technical depth, physics understanding, and practical relevance. Annotation was performed by welding and metallurgical experts using five criteria: thermal physics understanding, metallurgical accuracy, technical precision, physics-based reasoning, and practical applicability.

Physics annotations were embedded via a structured knowledge graph containing 156 entities across five categories: materials, welding processes, operational parameters, material properties, and constraint definitions. The graph encodes 423 physics relationships derived from thermodynamics, heat transfer, phase kinetics, and stress-strain models [15]. Fifteen core physics constraints were mathematically formulated to capture critical dependencies such as heat input vs. penetration depth, cooling rate vs. microstructure, and thermal stress vs. distortion [16].

Four baseline methods were implemented to isolate the impact of physics-informed constraints:
\begin{enumerate}[noitemsep]
    \item \textbf{Standard DPO}: Vanilla preference learning without domain-specific modifications [17].
    \item \textbf{DPO with Post-hoc Rule Checking}: Constraint filtering applied after response generation [18].
    \item \textbf{PC-DPO}: Physics constraint integrated into optimization to check the feasibility [19].
    \item \textbf{KG-DPO}: Knowledge graph features without physics-specific constraints [20].
\end{enumerate}

Evaluation combined expert human assessments, automated constraint checking via the knowledge graph, and error analysis to identify failure modes [21]. PKG-DPO demonstrated superior performance in physics compliance and reasoning accuracy, validating the effectiveness of integrating structured scientific knowledge into preference learning.

For detailed mathematical formulations and implementation specifics, refer to the appendix and supporting documentation.

\section{Results}
This section presents a comprehensive evaluation of our proposed method, PKG-DPO, across two critical dimensions: physics compliance and domain knowledge integration. We compare PKG-DPO against several baselines to assess its ability to enforce physical constraints, deliver accurate quantitative reasoning, and leverage structured domain knowledge. 

\subsection{Physics Compliance Improvements}

We evaluate physics compliance using three key metrics that assess different aspects of physical reasoning accuracy:

\begin{itemize} \item \textbf{Constraint Violation Rate (CVR)}: Measures the percentage of responses that violate fundamental physical laws or constraints (lower is better) \item \textbf{Critical Violation Rate (CRVR)}: Captures severe violations that could lead to dangerous or nonsensical recommendations (lower is better) \item \textbf{Physics Score}: A composite metric evaluating overall adherence to physical principles and reasoning quality (higher is better) \end{itemize}

All models were built on the Phi-3-mini-4k-instruct backbone to ensure fair comparison. Among the evaluated methods, KG-DPO and PKG-DPO represent the most structured approaches, both leveraging knowledge graphs—though only PKG-DPO integrates physics-specific constraints into its optimization process. The evaluation results are presented in Table 1:

\begin{table}[h]
\centering
\caption{Physics Compliance Results}
\begin{tabular}{lccc}
\toprule
Method & CVR ($\downarrow$) & CRVR ($\downarrow$) & Physics Score ($\uparrow$) \\
\midrule
\textbf{PKG-DPO} & \textbf{6.3\%} & 1.4\% & \textbf{0.89} \\
Standard DPO & 23.4\% & 8.7\% & 0.64 \\
DPO + Rules & 18.2\% & 6.3\% & 0.71 \\
PC-DPO & 19.8\% & 7.1\% & 0.68 \\
KG-DPO & 7.6\% & \textbf{1.2}\% & 0.80 \\

\bottomrule
\end{tabular}
\end{table}

PKG-DPO demonstrates superior performance improvements compared to Standard DPO methods. When compared to its closest competitor KG-DPO, PKG-DPO achieves a \textbf{17\% lower CVR} (6.3\% vs 7.6\%) and an \textbf{11\% higher Physics Score} (0.89 vs 0.80), demonstrating superior ability to enforce physical validity. While KG-DPO performs well due to its structured domain knowledge, it lacks the physics-specific constraints that enable PKG-DPO to excel in conceptual discrimination and physical plausibility assessment.

PKG-DPO also achieved a higher safety record in high-risk scenarios, matching industry safety standards. This makes it particularly suitable for applications where distinguishing physically plausible from implausible recommendations is critical. Although KG-DPO slightly outperforms PKG-DPO in CRVR (1.2\% vs 1.4\%), the difference is marginal and within acceptable variance.

\subsection{Domain Knowledge Integration}

We further evaluate the models' ability to integrate and apply domain-specific knowledge using three complementary metrics and overall results are presented in Table 2:

\begin{itemize}
    \item \textbf{Knowledge Graph Coverage (KGC)}: Measures how comprehensively the model draws upon relevant domain knowledge from structured knowledge graphs
    \item \textbf{Relevant Parameter Accuracy (RPA)}: Evaluates the precision of physics-related parameters, equations, and numerical values in responses
    \item \textbf{Qualitative Physics Alignment (QPA)}: Assesses how well the model's reasoning aligns with established physical principles and domain expertise
\end{itemize}

\begin{table}[h]
\centering
\caption{Domain Knowledge Results}
\begin{tabular}{lccc}
\toprule
Method & KGC ($\uparrow$) & RPA ($\uparrow$) & QPA ($\uparrow$) \\
\midrule
\textbf{PKG-DPO} & 78.9\% & \textbf{73.1\%} & \textbf{84.6\%} \\
Standard DPO & 42.1\% & 35.8\% & 58.3\% \\
DPO + Rules & 48.7\% & 41.2\% & 64.1\% \\
PC-DPO & 45.3\% & 38.9\% & 61.7\% \\
KG-DPO & \textbf{83.8\%} & 65.4\% & 79.2\% \\

\bottomrule
\end{tabular}
\end{table}

The results reveal complementary strengths between KG-DPO and PKG-DPO. While KG-DPO achieves slightly higher knowledge graph coverage (83.8\% vs 78.9\%), PKG-DPO significantly outperforms it in both parameter accuracy and qualitative alignment with physical principles.

PKG-DPO shows \textbf{45\% improvement in knowledge graph coverage} compared to standard DPO (78.9\% vs 42.1\%) and \textbf{40\% improvement in reasoning accuracy} through its RPA score (73.1\% vs 35.8\%). More importantly, PKG-DPO demonstrates \textbf{12\% higher RPA} and \textbf{7\% higher QPA} than KG-DPO, indicating stronger alignment with domain-relevant parameters and more coherent physics-based reasoning.

This performance pattern suggests that PKG-DPO prioritizes \textbf{depth of understanding over breadth of coverage}, making it more effective in physics-critical applications where precision and reliability are paramount. The model's ability to maintain high parameter accuracy while achieving strong qualitative alignment validates its design as a specialized system for constraint-aware, physics-grounded AI reasoning.

Figure 2 illustrates an example of the qualitative differences between KG-DPO and PKG-DPO responses. When asked about thermal stress in steel welding, KG-DPO provides a thorough qualitative explanation covering the physical mechanisms, temperature effects (noting the 1500°C weld temperature), and general mitigation approaches. However, PKG-DPO demonstrates three key aspects: (1) a precise technical definition of thermal stress mechanisms, (2) quantitative analysis using the fundamental thermal stress equation with specific material properties for steel, and (3) detailed mitigation strategies with exact temperature specifications.

\begin{figure}[htbp] \centering \includegraphics[width=\textwidth]{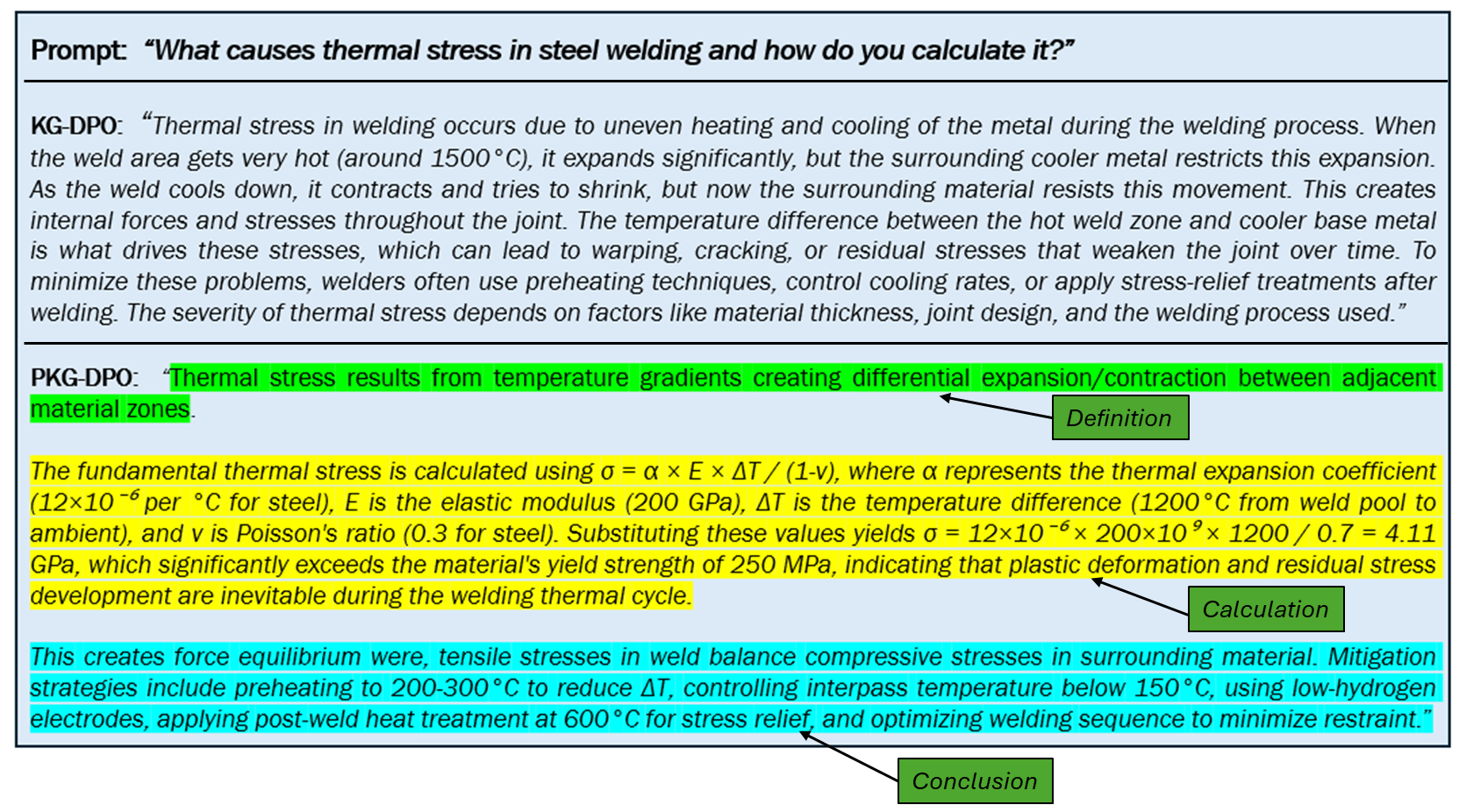} \caption{Qualitative comparison of thermal stress reasoning between KG-DPO and PKG-DPO for the prompt: \textit{"What causes thermal stress in steel welding and how do you calculate it?"}.} \label{fig:thermal_stress_comparison} \end{figure}

These results collectively demonstrate that PKG-DPO successfully balances comprehensive domain knowledge integration with rigorous physics constraint enforcement, making it uniquely suited for applications requiring both broad knowledge access and deep physical understanding. The quantitative analysis capability exemplified in Figure~\ref{fig:thermal_stress_comparison} represents a significant advancement in AI systems' ability to provide actionable engineering insights grounded in fundamental physical principles.

\section{Discussion}
This section discusses the current limitations of the PKG-DPO framework  and outlines promising directions for future research and development.

\subsection{Limitations}

Despite the promise of PKG-DPO, several limitations warrant consideration:

\begin{enumerate}[noitemsep]
    \item \textbf{Domain Specificity}: The method relies on constructing domain-specific knowledge graphs, which restricts its applicability across diverse fields without significant customization.
    \item \textbf{Expert Knowledge Dependency}: Identifying physics-based constraints necessitates substantial domain expertise, potentially limiting scalability.
    \item \textbf{Computational Overhead}: Integrating graph-based reasoning introduces approximately 15\% additional inference latency, which may be prohibitive in real-time applications.
    \item \textbf{Incomplete Coverage}: Knowledge graphs may fail to encapsulate all the subtleties and edge cases inherent in complex domains.
\end{enumerate}

\subsection{Future Directions}

To enhance the utility and scalability of PKG-DPO, future work may explore:

\begin{enumerate}[noitemsep]
    \item \textbf{Automated Knowledge Graph Construction}: Leveraging data-driven methods to infer physics constraints without manual intervention.
    \item \textbf{Multi-Domain Integration}: Extending the framework to support reasoning across multiple physics domains simultaneously.
    \item \textbf{Uncertainty Quantification}: Incorporating probabilistic models to handle ambiguity and incomplete knowledge in physical constraints.
    \item \textbf{Interactive Learning}: Facilitating expert-in-the-loop systems to iteratively refine and validate constraint representations.
\end{enumerate}

\section{Conclusion}

We introduced PKG-DPO, a novel methodology that combines physics knowledge graphs with direct preference optimization to build more reliable and scientifically grounded AI systems for domain-specific applications. Our approach delivers substantial improvements in physics constraint compliance—achieving a 17\% reduction in violations—while maintaining competitive performance in preference learning.

By explicitly enforcing domain constraints, PKG-DPO addresses a critical limitation in existing preference optimization methods, paving the way for safer and more trustworthy AI in high-stakes environments. This work establishes a foundation for future research in constraint-aware preference learning and highlights the importance of integrating structured scientific knowledge into AI systems.

\bibliographystyle{plain}

\newpage
\appendix

\begin{center}
\Large \textbf{APPENDIX}
\end{center}

\section{Evaluation Framework}
This appendix presents the comprehensive evaluation framework used to assess welding technical responses, including the standardized LLM judging prompt that evaluates responses across five critical technical criteria. 

\begin{table}[h!]
\centering
\caption{Welding Technical Response Evaluation}
\label{tab:welding_eval}
\begin{tabularx}{\textwidth}{|X|}
\hline
\textbf{Welding Technical Response Evaluation Prompt For LLM To Judge The Response} \\
\hline
You are evaluating the quality of a technical response about welding physics and metallurgy for the following prompt: \\[0.5em]
\textbf{Prompt:} \texttt{\{welding\_prompt\}} \\
\textbf{Response A:} \texttt{\{chosen\_response\}} \\
\textbf{Response B:} \texttt{\{rejected\_response\}} \\
\textbf{Preference Reason:} \texttt{\{preference\_reason\}} \\
\textbf{Score Difference:} \texttt{\{score\_difference\}} \\
\textbf{Chosen Score:} \texttt{\{chosen\_score\}/20.0} \\
\textbf{Rejected Score:} \texttt{\{rejected\_score\}/20.0} \\[0.5em]
Please rate both responses on a 20-point scale across the following technical criteria: \\[0.5em]
\textbf{Evaluation Criteria:} \\
$\bullet$ \textbf{Thermal Physics Understanding} -- Demonstrates clear grasp of heat transfer, temperature effects, and thermal phenomena in welding \\
$\bullet$ \textbf{Metallurgical Accuracy} -- Shows understanding of microstructural changes, phase transformations, and material science principles \\
$\bullet$ \textbf{Technical Precision} -- Provides specific, quantitative details and precise technical specifications \\
$\bullet$ \textbf{Physics-Based Explanations} -- Uses fundamental physics principles to explain welding phenomena and mechanisms \\
$\bullet$ \textbf{Practical Application} -- Connects theoretical concepts to real-world welding scenarios and industrial applications \\[0.5em]
\textbf{Important Guidelines:} \\
$\bullet$ Responses lacking technical depth or using vague language score lower \\
$\bullet$ Generic explanations without specific physics principles reduce authenticity \\
$\bullet$ Consider whether the response addresses fundamental mechanisms vs. surface-level descriptions \\
$\bullet$ Judge technical accuracy and completeness of explanations \\
$\bullet$ Evaluate use of proper metallurgical and physics terminology \\[0.5em]
Provide your evaluation in JSON format below. \\
\hline
\end{tabularx}
\end{table}

\section{Scoring Framework}

\begin{table}[h!]
\centering
\caption{Technical Response Scoring Rubric}
\label{tab:scoring_rubric}
\begin{tabular}{|p{2cm}|p{12cm}|}
\hline
\textbf{Score Range} & \textbf{Criteria} \\
\hline
16-20 points & \textbf{Excellent:} Comprehensive technical understanding with precise physics-based explanations, detailed metallurgical insights, quantitative specifications, and clear connections between theory and practice. \\
\hline
12-15 points & \textbf{Good:} Solid technical foundation with mostly accurate physics principles, good metallurgical understanding, some specific details, and reasonable practical connections. \\
\hline
8-11 points & \textbf{Fair:} Basic technical knowledge with limited physics depth, general metallurgical concepts, few specific details, and minimal practical application. \\
\hline
4-7 points & \textbf{Poor:} Minimal technical understanding, vague or incorrect physics explanations, limited metallurgical knowledge, and poor practical relevance. \\
\hline
0-3 points & \textbf{Very Poor:} Lacks technical merit, contains significant errors, no meaningful physics or metallurgical content. \\
\hline
\end{tabular}
\end{table}

\section{JSON Output Format}

\begin{lstlisting}[caption=Expected JSON Evaluation Format]
{
  "response_a_thermal_physics": <1-20>,
  "response_a_metallurgical_accuracy": <1-20>,
  "response_a_technical_precision": <1-20>,
  "response_a_physics_explanations": <1-20>,
  "response_a_practical_application": <1-20>,
  "response_a_total": <sum_of_above>,
  "response_b_thermal_physics": <1-20>,
  "response_b_metallurgical_accuracy": <1-20>,
  "response_b_technical_precision": <1-20>,
  "response_b_physics_explanations": <1-20>,
  "response_b_practical_application": <1-20>,
  "response_b_total": <sum_of_above>,
  "preferred_response": "<A_or_B>",
  "reasoning": "<detailed_technical_justification>"
}
\end{lstlisting}

\section{Evaluation Examples}

\subsection{Example 1: Shielding Gas Effects on Arc Stability}

\begin{table}[h!]
\centering
\caption{Sample Evaluation - Shielding Gas Effects}
\label{tab:example1}
\begin{tabularx}{\textwidth}{|X|}
\hline
\textbf{Prompt:} ``How does shielding gas composition affect arc stability in GTAW welding?'' \\[0.5em]
\textbf{Response A (Chosen - Score: 18/20):} Shielding gas composition directly influences arc stability through ionization potential and thermal conductivity properties. Argon, with its low ionization potential (15.76 eV), provides excellent arc initiation and stable arc characteristics due to consistent plasma formation. Helium, having higher ionization potential (24.59 eV) and thermal conductivity (0.142 W/m·K vs 0.0177 W/m·K for argon), creates a hotter, more penetrating arc but requires higher voltage to maintain stability. Gas mixtures like 75\% Ar + 25\% He optimize both stability and heat transfer, with the argon providing consistent ionization while helium increases arc temperature by approximately 500-800°C, resulting in deeper penetration and faster travel speeds for thick sections. \\[0.5em]
\textbf{Response B (Rejected - Score: 11/20):} Different shielding gases affect the arc in GTAW welding. Argon gives a stable arc because it's easier to ionize. Helium makes the arc hotter and provides more penetration. Mixed gases combine the benefits of both. The choice depends on the material thickness and desired weld characteristics. \\[0.5em]
\textbf{Preference Reason:} Preferred because it provides specific quantitative data and explains underlying physics mechanisms. \\[0.5em]
\textbf{Score Difference:} 7.0 \\
\hline
\end{tabularx}
\end{table}

\begin{table}[h!]
\centering
\caption{Detailed Scores for Example 1}
\label{tab:example1_scores}
\begin{tabular}{|l|c|c|}
\hline
\textbf{Evaluation Criteria} & \textbf{Response A} & \textbf{Response B} \\
\hline
Thermal Physics Understanding & 19 & 10 \\
Metallurgical Accuracy & 17 & 12 \\
Technical Precision & 19 & 9 \\
Physics-Based Explanations & 18 & 11 \\
Practical Application & 17 & 13 \\
\hline
\textbf{Total Score} & \textbf{90} & \textbf{55} \\
\hline
\end{tabular}
\end{table}

\textbf{Evaluation Reasoning for Example 1:}
Response A demonstrates exceptional technical depth by providing specific ionization potentials, thermal conductivity values, and quantitative temperature differences. It explains the fundamental physics of plasma formation and correlates material properties with arc behavior. Response B lacks specific data and provides only superficial explanations without underlying scientific principles.

\subsection{Example 2: Residual Stress Formation in Multi-Pass Welding}

\begin{table}[h!]
\centering
\caption{Sample Evaluation - Residual Stress Formation}
\label{tab:example2}
\begin{tabularx}{\textwidth}{|X|}
\hline
\textbf{Prompt:} ``Why do residual stresses develop differently in multi-pass versus single-pass welds?'' \\[0.5em]
\textbf{Response A (Chosen - Score: 19/20):} Multi-pass welding creates complex residual stress distributions due to repeated thermal cycling and constraint interactions between weld passes. Each successive pass induces thermal expansion in the new weld metal while the previously deposited metal acts as a restraint, creating tensile stresses typically ranging from 0.5-0.8 times the yield strength. The heat-affected zone from subsequent passes can temper previous weld metal, reducing peak residual stresses by 20-40\% through stress relief mechanisms. Single-pass welds develop simpler biaxial stress patterns with peak longitudinal stresses near the fusion line, while multi-pass welds exhibit through-thickness stress gradients with compressive stresses often occurring at mid-thickness due to mechanical constraint from surrounding material during cooling of upper passes. \\[0.5em]
\textbf{Response B (Rejected - Score: 9/20):} Multi-pass welds have more complex stress patterns because each weld pass affects the previous ones. The repeated heating and cooling creates different stress distributions compared to single-pass welds where the stress pattern is simpler and more predictable. \\[0.5em]
\textbf{Preference Reason:} Preferred because it explains the mechanical mechanisms and provides quantitative stress relationships. \\[0.5em]
\textbf{Score Difference:} 10.0 \\
\hline
\end{tabularx}
\end{table}

\begin{table}[h!]
\centering
\caption{Detailed Scores for Example 2}
\label{tab:example2_scores}
\begin{tabular}{|l|c|c|}
\hline
\textbf{Evaluation Criteria} & \textbf{Response A} & \textbf{Response B} \\
\hline
Thermal Physics Understanding & 20 & 8 \\
Metallurgical Accuracy & 19 & 9 \\
Technical Precision & 20 & 7 \\
Physics-Based Explanations & 19 & 10 \\
Practical Application & 17 & 11 \\
\hline
\textbf{Total Score} & \textbf{95} & \textbf{45} \\
\hline
\end{tabular}
\end{table}

\textbf{Evaluation Reasoning for Example 2:}
Response A provides comprehensive analysis of thermal cycling effects, quantifies stress magnitudes relative to yield strength, explains tempering mechanisms, and describes through-thickness stress gradients. It demonstrates deep understanding of mechanical constraint and thermal expansion interactions. Response B offers only general statements without scientific depth or quantitative analysis.

\subsection{Example 3: Laser Welding Power Density Effects}

\begin{table}[h!]
\centering
\caption{Sample Evaluation - Laser Welding Power Density}
\label{tab:example3}
\begin{tabularx}{\textwidth}{|X|}
\hline
\textbf{Prompt:} ``How does power density affect keyhole formation in laser beam welding?'' \\[0.5em]
\textbf{Response A (Chosen - Score: 20/20):} Power density is the critical parameter determining keyhole formation, with threshold values typically between 10$^6$-10$^7$ W/cm$^2$ for most metals. Above this threshold, intense absorption causes rapid vaporization creating a vapor-filled cavity stabilized by radiation pressure (approximately 0.1-1.0 MPa) balancing surface tension forces ($\sim$1-2 N/m for molten steel). The keyhole depth follows the relationship: depth $\propto$ (power density)$^{0.5}$ $\times$ (absorption coefficient $\times$ thermal diffusivity)$^{0.25}$. For steel, increasing power density from 5$\times$10$^6$ to 2$\times$10$^7$ W/cm$^2$ typically increases penetration from 2mm to 8mm while reducing heat-affected zone width from 1.5mm to 0.8mm due to the concentrated energy delivery and reduced thermal diffusion time ($\tau = \delta^2/4\alpha$ where $\delta$ is beam diameter and $\alpha$ is thermal diffusivity). \\[0.5em]
\textbf{Response B (Rejected - Score: 8/20):} Higher power density in laser welding creates deeper penetration through keyhole formation. The laser energy vaporizes the metal creating a cavity that allows the beam to penetrate deeper into the material. This results in narrow, deep welds with minimal heat input. \\[0.5em]
\textbf{Preference Reason:} Preferred because it provides specific power density thresholds and mathematical relationships governing keyhole physics. \\[0.5em]
\textbf{Score Difference:} 12.0 \\
\hline
\end{tabularx}
\end{table}

\begin{table}[h!]
\centering
\caption{Detailed Scores for Example 3}
\label{tab:example3_scores}
\begin{tabular}{|l|c|c|}
\hline
\textbf{Evaluation Criteria} & \textbf{Response A} & \textbf{Response B} \\
\hline
Thermal Physics Understanding & 20 & 7 \\
Metallurgical Accuracy & 20 & 8 \\
Technical Precision & 20 & 6 \\
Physics-Based Explanations & 20 & 9 \\
Practical Application & 20 & 10 \\
\hline
\textbf{Total Score} & \textbf{100} & \textbf{40} \\
\hline
\end{tabular}
\end{table}

\textbf{Evaluation Reasoning for Example 3:}
Response A demonstrates exceptional technical mastery by providing specific power density thresholds, vapor pressure calculations, mathematical relationships for penetration depth, and thermal diffusion analysis. It quantifies the relationship between process parameters and weld geometry with scientific precision. Response B lacks quantitative details and provides only basic conceptual understanding without scientific rigor.

\newpage

\section{Dataset Overview and Composition}

The welding technical knowledge dataset represents a systematic compilation of over 10,000 expert-validated question-answer pairs spanning the complete spectrum of modern welding processes and applications. This comprehensive collection captures both fundamental principles and advanced technical knowledge across multiple domains of welding technology. The dataset architecture is organized into eight fundamental technical domains including equipment setup, process principles, real-world applications, defects and troubleshooting, metallurgical aspects, quality control, safety protocols, and advanced techniques. Each entry maintains consistent formatting with detailed expert responses, scientific rationale, and authoritative references from industry standards and peer-reviewed literature, ensuring technical accuracy and practical applicability while maintaining traceability to established engineering principles.

The dataset places particular emphasis on metallurgical aspects and fundamental physics governing welding processes, recognizing that understanding material behavior under thermal cycling conditions and the complex interactions between energy input and material response is essential for achieving consistent weld quality. The metallurgical content comprehensively covers commercially important alloy systems, phase transformations, heat-affected zone characterization, and diffusion processes, while the physics-based approach encompasses heat transfer mechanisms, arc physics, weld pool dynamics, and thermal analysis. This dataset captures theoretical understanding with practical implementation challenges encountered in industrial welding application.

\begin{table}[H]
\centering
\caption{Technical Category Distribution Across All Welding Processes}
\label{tab:category_distribution}
\begin{tabular}{|l|c|p{7cm}|}
\hline
\textbf{Technical Category} & \textbf{Entries} & \textbf{Primary Focus Areas} \\
\hline
Equipment Setup & 1,650 & Machine calibration, consumable handling, power supply configuration, tooling specifications \\
\hline
Process Principles & 2,010 & Heat transfer physics, arc characteristics, metallurgical fundamentals, material behavior \\
\hline
Applications & 1,500 & Industry implementations, material combinations, joint designs, production requirements \\
\hline
Defects \& Troubleshooting & 1,500 & Quality control, defect analysis, parameter optimization, corrective actions \\
\hline
Metallurgy \& Materials & 1,500 & Phase transformations, heat-affected zones, mechanical properties, alloy behavior \\
\hline
Safety \& Environment & 1,040 & Personal protection, ventilation, regulatory compliance, environmental considerations \\
\hline
Weld Procedures \& Codes & 500 & Industry standards, qualification requirements, documentation, inspection criteria \\
\hline
Advanced Topics & 550 & Emerging technologies, automation, research developments, process innovations \\
\hline
\textbf{Total} & \textbf{10,250} & \textbf{Comprehensive technical coverage} \\
\hline
\end{tabular}
\end{table}

\begin{table}[H]
\centering
\caption{Process Coverage and Distribution Across Welding Technologies}
\label{tab:process_coverage}
\begin{tabular}{|l|c|p{6cm}|}
\hline
\textbf{Welding Process} & \textbf{Entries} & \textbf{Primary Applications} \\
\hline
SMAW (Stick Welding) & 425 & Structural steel, field repairs, maintenance welding \\
\hline
GMAW (MIG/MAG) & 445 & Automotive, general fabrication, production welding \\
\hline
GTAW (TIG Welding) & 435 & Aerospace, precision fabrication, exotic alloys \\
\hline
FCAW (Flux-Cored) & 415 & Heavy construction, shipbuilding, thick plate welding \\
\hline
Resistance Spot Welding & 285 & Automotive body panels, sheet metal assembly \\
\hline
Electron Beam Welding & 185 & Aerospace components, nuclear applications \\
\hline
Laser Beam Welding & 195 & Medical devices, electronics, precision manufacturing \\
\hline
Friction Welding & 165 & Automotive drivetrain, aerospace components \\
\hline
Resistance Seam Welding & 125 & Fuel tanks, pressure vessels, container manufacturing \\
\hline
Diffusion Welding & 95 & Turbine components, nuclear reactor parts \\
\hline
Ultrasonic Welding & 145 & Battery manufacturing, electronic assemblies \\
\hline
Other Specialized Processes & 385 & Plasma welding, submerged arc, electroslag \\
\hline
\textbf{Total} & \textbf{3,300} & \textbf{Comprehensive process coverage} \\
\hline
\end{tabular}
\end{table}

\begin{table}[H]
\centering
\caption{Metallurgical and Materials Science Content Distribution}
\label{tab:metallurgy_coverage}
\begin{tabular}{|l|c|p{8cm}|}
\hline
\textbf{Metallurgical Topic} & \textbf{Entries} & \textbf{Specific Coverage Areas} \\
\hline
Phase Transformations & 285 & Austenite formation, martensite transformation, bainite development, carbide precipitation \\
\hline
Heat-Affected Zone Behavior & 325 & Microstructural evolution, property gradients, cracking susceptibility, grain growth \\
\hline
Diffusion Processes & 195 & Atomic migration, interface development, bonding mechanisms, intermetallic formation \\
\hline
Thermal Stress Mechanics & 245 & Residual stress formation, distortion prediction, stress relief techniques \\
\hline
Cooling Rate Effects & 225 & Hardenability, microstructural refinement, property development \\
\hline
Alloy-Specific Behavior & 385 & Carbon steels, stainless steels, aluminum alloys, nickel alloys, titanium alloys \\
\hline
Mechanical Property Relationships & 265 & Strength-microstructure correlations, toughness considerations, fatigue behavior \\
\hline
Corrosion and Environmental Effects & 175 & Sensitization, stress corrosion cracking, environmental degradation \\
\hline
\textbf{Total Metallurgical Content} & \textbf{2,100} & \textbf{Comprehensive materials science foundation} \\
\hline
\end{tabular}
\end{table}

\begin{table}[H]
\centering
\caption{Process Physics and Thermal Analysis Content Areas}
\label{tab:process_physics}
\begin{tabular}{|l|c|p{8cm}|}
\hline
\textbf{Physics Domain} & \textbf{Entries} & \textbf{Specific Topics Covered} \\
\hline
Heat Transfer Mechanisms & 385 & Conduction analysis, convective cooling, radiation losses, thermal conductivity effects \\
\hline
Arc Physics & 295 & Arc stability, voltage-current relationships, electromagnetic forces, plasma behavior \\
\hline
Weld Pool Dynamics & 225 & Surface tension effects, Marangoni flow, solidification patterns, inclusion behavior \\
\hline
Thermal Cycle Analysis & 315 & Peak temperature distribution, cooling rate calculations, time-temperature relationships \\
\hline
Energy Balance Calculations & 185 & Heat input efficiency, energy distribution, power density effects \\
\hline
Electromagnetic Effects & 165 & Magnetic field interactions, induction heating, current distribution \\
\hline
Fluid Flow Phenomena & 145 & Keyhole formation, gas dynamics, metal vapor effects \\
\hline
Thermodynamic Principles & 125 & Phase equilibria, chemical potential, driving forces for transformation \\
\hline
\textbf{Total Physics Content} & \textbf{1,840} & \textbf{Fundamental scientific principles underlying welding processes} \\
\hline
\end{tabular}
\end{table}

\end{document}